# A Cookbook for Community-driven Data Collection of Impaired Speech in Low-Resource Languages

*Sumaya Ahmed Salihs[1], Isaac Wiafe[1], Jamal-Deen Abdulai[1], Elikem Doe Atsakpo[2], Gifty Ayoka[3], Richard Cave[4,5], Akon Obu Ekpezu[6], Catherine Holloway[4,5], Katrin Tomanek[7], Fiifi Baffoe Payin Winful[1]*

[1]Department of Computer Science, University of Ghana, Ghana
[2]School of Computing and Engineering, University of West London, United Kingdom
[3]Talking Tipps Africa Foundation, Ghana
[4]University College London Interaction Centre, United Kingdom
[5]Global Disability Innovation Hub, University College London, United Kingdom
[6]Department of Information Processing Science, University of Oulu, Finland
[7]Google Research, USA

```
sasalihs@st.ug.edu.gh, iwiafe@ug.edu.gh, JAbdulai@ug.edu.gh,
33136323@student.uwl.ac.uk, giphty2001@gmail.com, richard.cave.17@ucl.ac.uk,
akon.ekpezu@oulu.fi, C.holloway@ucl.ac.uk, katryn.tomanek@gmail.com,
fbpwinful@st.ug.edu.gh
```

## Abstract

This study presents an approach for collecting speech samples to build Automatic Speech Recognition (ASR) models for impaired speech, particularly, low-resource languages. It aims to democratize ASR technology and data collection by developing a "cookbook" of best practices and training for community-driven data collection and ASR model building. As a proof-of-concept, this study curated the first open-source dataset of impaired speech in Akan: a widely spoken indigenous language in Ghana. The study involved participants from diverse backgrounds with speech impairments. The resulting dataset, along with the cookbook and open-source tools, are publicly available to enable researchers and practitioners to create inclusive ASR technologies tailored to the unique needs of speech impaired individuals. In addition, this study presents the initial results of finetuning open-source ASR models to better recognize impaired speech in Akan.

## 1. Introduction

Automatic Speech Recognition (ASR) technology has transformed human-human and human-computer communications. It facilitates understanding through real-time speech captioning [1], [2] and supports hands-free computing (e.g. email dictations, emails, online information retrieval, and automatic language translation). ASR is used to control smart home activities, such as changing television channels, heating, ventilation, air conditioning, and adjusting lighting. Although it continues to be useful, most of these technologies do not cater to speech diversity and are often optimized for 'standard' or typical speech. Therefore, they fail to benefit individuals with impaired speech such as dysarthria, stammering, or cleft palate who often experience reduced ASR accuracy. Prior studies have demonstrated the potential benefits of speech recognition technologies in English for distinct impaired speech[2], [3], [4]. While this benefits English speakers, it is imperative to extend similar technologies to low-resource languages (LRLs). LRL communities have limited access to assistive technologies and speech and language therapy (SLT) services [5], [6], [7]. Hence, the availability of ASR technologies in LRLs will facilitate effective communication for those with speech impairments, especially in sub-Saharan Africa, where there are insufficient speech therapy resources [7]. This study is part of a larger initiative that seeks to collect, validate, and create a large corpus of impaired speech in LRLs. It reports the findings of a pilot study in the Akan language from Ghana, by discussing the methods, challenges, and lessons learned from the data collection, validation, and testing of the dataset to adapt ASR models.

## 2. Related work

Between 28% and 49% of individuals worldwide encounter communication difficulties [8]. Individuals with Amyotrophic Lateral Sclerosis or Motor Neuron Disease experience progressive physical and speech changes, which restrict their ability to engage with their environment [9]. Approximately 65% of individuals with Traumatic Brain Injury suffer from speech impairment [10], and more than 70% of individuals with Parkinson's disease experience significant speech-related challenges [11]. These are a few examples from a much longer list of conditions or situations that significantly affect an individual's speech. Yet, current ASR models fail to accurately understand many of these speakers [12]. Advocates of these communities emphasize that ASR can neither be considered inclusive nor used pervasively by communities that need it until it effectively serves the needs of all individuals, including those with impaired speech. Although a few commercially available ASR models have been adapted in limited ways for English speakers with impaired speech, none are adapted for LRLs. The lack of functional ASR technologies for impaired speech in LRLs may be attributed to the lack of relevant data for the development of such technologies. Currently, existing impaired speech datasets such as UA-Speech [13], TORGO [14] and Google's Euphonia [1] are mostly in English. Moreso, some [1] are not openly accessible. Although English is the official language in many LRL countries including Ghana, studies show that ASR models trained in American English have low caption accuracy when applied to Ghanaian English [3]. Also, these models do not address the linguistic needs of the local population with impaired speech who may not speak English. This suggests a need for an impaired speech corpus for LRLs.

## 3. Methods

To accommodate the linguistic diversity and communication needs of these underserved populations, a community-driven data-collection approach was used. We selected a target language (i.e. Akan), and speech conditions namely cerebral palsy, cleft palate, and stammering for the pilot study. Akan is the most widely spoken indigenous language in Ghana, with over nine million speakers [15]. Selecting a more prevalent indigenous language allowed for a diverse participant pool and enhanced the applicability of the data with substantial cultural and economic relevance. Akan is rich in linguistic diversity and provides a robust foundation for understanding the challenges associated with project scaling. We engaged local stakeholders, including linguists, speech and language therapists (SLTs), and community organizations in participant recruitment and data collection. The collaborative approach in participant recruitment and data collection sought to foster trust and encourage community participation. This is critical for long-term project success.

### 3.1. Data Collection Tool and Participant Recruitment

We adopted an open-source Android mobile application "UGSpeechData App", which was originally designed to collect audio recordings from standard speakers of Ghanaian languages on the UGSpeechData project [16] using image prompts, and fused it with methods used for collecting speech impaired data in English for Project Euphonia [1]. The UGSpeechData project used image prompts cutting across 50 culturally relevant themes whereas Project Euphonia used prompted speech. For this study, the UGSpeechData App was updated to include 1000 image prompts and 1000 text prompts. The text prompts were generated from a selection of 1000 images and described by four Akan linguists (two males and two females). The descriptions of the images were either short or long sentences. Short sentences were typically made of three to five words, while long sentences were made of five to ten words. To support participants who are unable to read the Akan language, we added a voice recording of the text prompts. Participants listened to and repeated the prompts for recording. Participants were recruited using convenience sampling from Kumasi, Accra Akosombo, and Koforidua, representing urban and rural settings spanning a range of regions across Ghana. SLTs facilitated participant selection based on intelligibility scoring of sentence-level speech by [17]. The demographic characteristics of the participants are shown in Table 1.

Table 1: *Participant Demographics*

| Etiology | N | Females | Males | Total recordings |
|---|---|---|---|---|
| Cerebral Palsy | 53 | 24 | 29 | 8050 |
| Stammering | 3 | 0 | 3 | 1350 |
| Cleft | 1 | 0 | 1 | 364 |
| Total | 57 | 24 | 33 | 9764 |

### 3.2. Facilitating Recordings for Impaired Speech

Data was collected using in-person and virtual methods. For the in-person method, participants were invited to a quiet facility near their residence, school, or care facility for recording. Some were invited in groups and others individually. To prioritize who to record, we measured the speech impairment severity and cognitive ability. We selected participants with impaired speech assessed as mildly impaired or more severe by the SLTs. All participants were debriefed on the project's aim and trained to use the app for specific purposes, including logging in to the app, accepting informed consent, updating their personal information, and recording. Samsung Galaxy Tab A8 10.5 (2021) was used for in-person recordings whereas participants used their own devices for online participation, except for those who requested a device.

Upon accepting to participate and signing the informed consent form on the app, each participant was assigned 300 images and 150 text prompts. For image prompts, the app displayed a sequence of images. The participants described the image in their own words but followed a predefined set of rules. For text prompts, a sequence of short to long sentences in Akan was displayed on the screen for participants to read. Those who were unable to read used the pre-recorded audio of the displayed text prompt. Each participant was required to (i) verify the personal information displayed (ii) sign the consent form (iii) record in Akan as they typically speak and use English for loanwords only (iv) provide between two to sixty seconds descriptions of what they see in the displayed images, and (v) read the displayed text or repeat what they heard verbatim. To avoid biasing the dataset with similar words and phrases, participants were encouraged to avoid words such as mfoni no (meaning "the image"), mehu ("I see"), wɔ mfoni no mu ("in this picture"). Participants were also instructed to avoid filler sounds like errr, errmmm, hmmm, mmm (addictive), or any form of profanity. Where necessary, facilitators (including their caregivers) supported the participants in the ideation of image descriptions. Although recording sessions for most of the etiologies were 2 hours, participants with cerebral palsy spent about 4 hours to produce 30 minutes of recordings including intermediate breaks. Online participants included individuals who stammer or with cleft palate. Similar rules and procedures were explained to them via video conferencing. To ensure data quality, the initial ten recordings for each online participant underwent quality check and approval before they were allowed to record more.

### 3.3. Validation, Quality Control, and Transcription

Nine professionally trained linguists performed quality checks and transcribed the collected audio samples following predefined transcription guidelines and rules adapted and modified from similar projects. Transcriptions were performed using a custom desktop transcription app. Considering that the rules for written Akan are evolving rapidly; there was the need for all linguists to agree on specific transcription rules. Particularly, how English words must be spelled during code-switching and how to deal with neologisms that are not formally recognized in the Akan language. Transcribers were asked to flag speech samples with issues such as excessive background noise, audible second voice, and inaudible recordings. A double-blinded transcription was performed on each speech sample. In situations where it was challenging to understand a recording, an SLT proficient in the Akan language reviewed the recording. Furthermore, the severity levels of the speech samples were evaluated on a speaker level by SLTs. Speech samples were classified as mild, moderate, severe, profound, and within normal limits (WNL).

## 4. ASR Modelling

We demonstrated the utility of the curated dataset for evaluating and adapting ASR models. We used Whisper [18] a popular open-source ASR system trained on a multilingual dataset of over 680,000 hours of speech.

### 4.1. Training an ASR Base Model for Akan

Whisper supports ninety-nine languages but not Akan. Hence, we trained an Akan base model for standard speech using approximately 100 hours of transcribed Akan speech samples from the UGSpeechData dataset [16] consisting of 18,787 speech samples. The dataset was partitioned into training (16,480 utterances), test (1,522 utterances), and dev (405 utterances) sets, ensuring no speaker overlap across splits and a balanced distribution of speakers with age and gender. Whisper is available in various sizes. In this experiment, the "small" variant (twelve layers, totaling 244M parameters) was used. Yoruba was used as the base language for fine-tuning, as out of all the ninety-nine languages supported in Whisper its acoustic properties and alphabet are most similar to Akan [19]. For fine-tuning, we used small batch sizes of thirty-two and a learning rate of 1e-5 (with a short warm-up phase and no decay). SpecAugment [20], a data augmentation technique designed to increase the input variability by masking portions of the spectrogram in the time and frequency dimensions, was added. The dev set was used to determine the best checkpoint, and the hold-out test set was used to evaluate the word error rate (WER) and character error rate (CER). The best-performing model achieved a WER of 0.293 and a CER of 0.100.

### 4.2. Adaptation to Impaired Speech Data

One thousand, six hundred and seventy-two (1,672) utterances were removed from the impaired speech dataset [21] after quality checks and the remaining eight thousand and ninety-two (8,092) utterances were divided into train, test, and dev sets. Similar to the data partition methods used for developing the ASR base model for Akan, the partitioning process ensured no speaker overlap. Also, samples with severity levels labeled as "profound" and "WNL" (typical speech) were excluded from the test and dev set as suggested by [22]. Table 2 shows the summary of the resulting dataset.

Table 2: *Dataset for impaired speech model training: number of speakers (utterances in brackets) per data split*

| Split | dev | test | train |
|---|---|---|---|
| WNL | - | - | 6 (1260) |
| Mild | 1 (56) | 3 (254) | 11 (2020) |
| Moderate | 5 (140) | 5 (415) | 9 (2151) |
| Severe | 2 (39) | 3 (250) | 5 (1044) |
| Profound | - | - | 7 (463) |
| Total | 8 (235) | 11 (919) | 38 (6938) |

The ASR base model for Akan achieved a median WER of 84.6% across all eleven speakers in the impaired speech test set. This is not surprising, given that previous research [4] has shown that ASR models that are trained without impaired speech data perform poorly. Accordingly, we further fine-tuned our Akan ASR model on the impaired speech data (training split), using the same fine-tuning recipe. Table 3 shows that adaptation significantly reduced WER for all speakers, with a median relative WER reduction of 21.7%. However, at the per-speaker level, the WER reduction ranged from as low as 7% up to 48%. Table 3 also shows the WER and WER reduction measured on the test set for both the Akan base model and the Akan model adapted to impaired speech. Further analysis revealed that data quality was the primary source of high WERs after adaptation for many speakers. For example, a manual error analysis of speaker K12 showed that samples from this speaker exhibit several transcription challenges, including frequent repetitions (with some words omitted in the transcriptions), instances of truncated words or phrases, and the 'correction' of words spoken in the transcription. Also, poor audio quality, such as variations in speech volume, occasional background noise, and the presence of nonspeech elements, were frequent.

Table 3: *Results of tuning the Akan Whisper on Akan impaired speech data*

| SP_ID | Severity | WER Ak_bm | WER Ak_am | WERR abs | WERR rel |
|---|---|---|---|---|---|
| K12 | Mild | 0.92 | 0.78 | 0.15 | 0.16 |
| K45 | Mild | 0.68 | 0.51 | 0.17 | 0.25 |
| K8 | Mild | 0.79 | 0.58 | 0.21 | 0.27 |
| K14 | Moderate | 0.85 | 0.74 | 0.11 | 0.13 |
| K28 | Moderate | 0.92 | 0.70 | 0.23 | 0.24 |
| K4 | Moderate | 0.91 | 0.47 | 0.44 | 0.48 |
| K44 | Moderate | 0.80 | 0.59 | 0.21 | 0.26 |
| K7 | Moderate | 0.91 | 0.85 | 0.06 | 0.07 |
| K24 | Severe | 0.91 | 0.70 | 0.21 | 0.23 |
| K39 | Severe | 0.81 | 0.68 | 0.13 | 0.16 |
| K48 | Severe | 0.83 | 0.71 | 0.12 | 0.14 |

SP_ID = Speaker ID; Ak_bm = Akan base model; Ak_am = Akan adapted model, WERR= word error rate reduction

## 5. Discussion

This study was a pilot initiative to create a corpus of impaired speech in Akan. The corpus consists of over 30 hours of audio samples of which 20 hours were transcribed. This section presents insights and lessons learned from improving future data collection efforts from similar populations.

### 5.1. Data Collection

This study used culturally relevant image and text prompts to facilitate the creation of a diverse, yet representative dataset. Although it was anticipated that text prompts would simplify the transcription process and reduce cost, since the post-facto transcription of disordered speech is particularly challenging and error-prone [1], image prompt was observed to be more effective in eliciting utterances. Many participants encountered challenges in accurately reading text prompts or repeating pre-recorded sentences verbatim. Particularly, those with severe cerebral palsy demonstrated associated cognitive challenges combined with reading difficulties. This posed significant challenges for linguists and language therapists during transcription as they encountered considerable difficulties in deciphering the recorded speech. Image prompts facilitated the use of familiar words and promoted engagement, leading to a faster and better collection of utterances. This finding is contrary to similar studies [23] conducted in English where text prompts proved efficient. Contrary to similar studies [1] that found online data collection methods to be effective, in-person data collection was more effective for working with individuals with speech impairments. It allowed for personalized assistance from facilitators, who led participants through the recording sections and addressed challenges in real-time. The online data collection approach faced significant limitations, including low participation rates, device incompatibility, limited technological literacy among participants, and lack of real-time support. While online methods may offer convenience for standard speakers, face-to-face interactions remain indispensable for ensuring high-quality data collection from individuals with speech impairment. The recruitment of participants through convenience sampling proved beneficial. Collaboration with SLTs and local institutions enabled the inclusion of individuals with diverse speech impairments including cerebral palsy, cleft palate, and stammering. However, initial challenges arose from overestimating participants' ability to independently use the data collection devices and apps. Many participants required assistance in handling devices, describing images, or reading text prompts. To address this challenge, one facilitator to one participant assignment was implemented for the cerebral palsy group. This adaptation was a critical step in the process. Training facilitators rather than participants ensured consistency and adherence to the recording protocols. Future studies should allocate sufficient resources to recruit and train facilitators who are proficient in the target language, digitally literate, and capable of addressing the unique needs of the participants.

### 5.2. Data Transcription

The transcription phase provided valuable insights into the challenges associated with transcription accuracy and consistency. A structured transcription process was adopted. Samples for specific speakers were allocated to specific transcribers. This allowed transcribers to be more familiar with the unique speech patterns of speakers and ultimately, improved transcription quality. The two-round process, using double-blind cross-validation, identified frequent transcription differences between transcribers, which is symptomatic of the challenge that LRLs such as Akan, have a still-developing writing system. The process also revealed the difficulties of transcribing lengthy recordings of impaired speech spoken at a fast rate combined with multiple repetitions, topic changes, and background noise. Some participants had cognitive impairments which occasionally resulted in non-standard syntax and unexpected semantics. In such a situation, both WER and meaning preservation assessments may be poor measurements for captioning success because many recordings had inconsistent and inaccurate transcriptions. Despite these challenges, the study successfully collected over 30 hours of audio recordings. This demonstrates the feasibility of inclusive data collection with proper planning and adaptability. Findings from this study demonstrate the relevance of participant-centered approaches, thoughtful prompt design, and robust technical infrastructure in creating high-quality datasets. These lessons will inform future initiatives and contribute meaningfully to the advancement of speech recognition technologies for underrepresented languages.

### 5.3. ASR Modelling

The findings from this study necessitate the need for tailored approaches that account for varying levels of impairment severity, and the development of larger and more diverse datasets. These elements are essential for advancing ASR technologies to effectively support individuals with impaired speech, especially LRLs. A word error rate reduction (WERR) of 21.7% is low when compared to previous studies which recorded 31% WERR on a similarly sized ASR model and 39% WERR on a significantly larger (2b parameters) ASR model for English language [22]. However, several factors might have accounted for the higher WERR in those studies. For instance, the base model (in English) used in [22] relatively performed better than what was used in this study (Akan standard dataset). Additionally, they used a much larger corpus of non-standard speech (almost 1m of training samples). Their samples were generated using prompted speech and shorter phrases, whereas samples used in this study spanned between two to 59 seconds from spontaneous speech and speakers. Despite the low WERR, we envisage that with large datasets and improved data collection and transcription methods, the WERR will improve. This will enable the potential for ASR technologies to improve accessibility and foster inclusivity in traditionally underserved communities.

## 6. Conclusions and Future Work

This study serves as a proof of concept for developing a corpus of impaired speech in Akan to support the advancement of ASR technology in LRLs. The resulting dataset provides a foundation for building systems that are inclusive for individuals with speech impairments. This study found that in-person data collection is essential for collecting data from individuals with speech impairment, particularly those with cerebral palsy. Also, it is imperative to use images that have simpler narratives to enhance participant engagement whilst ensuring that the data collection tools, recording environments, and support are tailored to address the unique needs of participants. Additionally, continued development of the rules to build consistency for LRL transcription is imperative. The potential of this dataset extends beyond its immediate use for ASR model development. It serves as a valuable resource for studying speech patterns in individuals with impairments and contributes to linguistics research, speech therapy, and assistive technology development. It will facilitate the development of systems that can recognize impaired speech patterns and serve as a critical step in addressing the limitations of ASR technologies, considering that these technologies mostly do not accommodate speech disabilities and underserved communities. The dataset as well as the proposed data collection procedure provide an opportunity for expansion to other LRLs and support the creation of culturally and linguistically relevant ASR technologies. And ultimately, this will promote digital language inclusion for underrepresented communities. This project is part of the important effort to ensure AI is open, inclusive, transparent, ethical, safe, secure, and trustworthy. The dataset is publicly available [21].

## 7. Ethics

Ethical approval for this study was obtained from the Ethics Committee for Basic and Applied Sciences (ECBAS), University of Ghana. All participants including recorders, caregivers, facilitators, SLTs, and transcribers were compensated for their respective contributions. Any personal identifiable data (PID) was removed from the recordings. The consent form specifies that data will be used for research purposes only, participation is voluntary, and participants can withdraw at any point. The samples collected will be used following the data policies and approvals obtained.

## 8. Acknowledgement

This research was funded by Google Research Ghana (Google Gift). AT2030 and UKaid provided travel support for the UCL team.